\documentclass{article}
\usepackage{proceedings}
\usepackage{apalike}
\usepackage[english]{babel}
\usepackage{amsmath,amssymb,latexsym}
\usepackage{mathrsfs}

\def\pre{\mathtt{pre}}
\def\jus{\mathtt{jus}}
\def\con{\mathtt{con}}

\def\p{\mathtt{p}}
\def\j{\mathtt{j}}

\def\c{\mathtt{c}}
\def\tstruct{\langle \alpha , \ac \rangle}

\def\tseq{\langle a_{j} \rangle}

\def\das{\langle a^{\p} , \{a_{1}^{\j},\ldots,a_{k}^{\j}\} , a^{\c} 
\rangle}

\newcommand{\dt}{\langle \Sigma , \Delta \rangle}
\newcommand{\ag}{\langle \Xi , A \rangle}

\def\CCA{{\rm T}_{\!A}}

\def\ia{\mathrm{I}}

\def\SP{\mathrm{sp}}

\def\ac{\mathcal{A}}

\def\ls{\mathscr{L}}

\def\var{\mathrm{Var}}

\newtheorem{thm}{Theorem}

\newtheorem{defn}{Definition}

\newtheorem{cor}{Corollary}
\newtheorem{exmp}{Example}

\title{Embedding Default Logic in Propositional Argumentation 
Systems}
\author{{\bf Dritan Berzati}\\
Department of Informatics\\
University 
of Fribourg\\
rue de Faucigny 2\\
1700 Fribourg, Switzerland\\
{\small \texttt{dritan.berzati@unifr.ch}}\\
\And
{\bf Bernhard 
Anrig}\\
Department of Informatics\\
University of Fribourg\\
rue de 
Faucigny 2\\
1700 Fribourg, Switzerland\\
{\small \texttt{bernhard.anrig@unifr.ch}}\\
\And
{\bf J\"urg Kohlas}\\
Department of Informatics\\
University of Fribourg\\
rue de Faucigny 
2\\
1700 Fribourg, Switzerland\\
{\small 
\texttt{juerg.kohlas@unifr.ch}}
}

\begin{document}

\maketitle

\begin{abstract}
 In this paper we present a transformation of finite propositional 
default 
theories
 into so-called propositional argumentation systems.  This 
transformation
 allows to characterize all notions of Reiter's default logic in the
 framework of argumentation systems. As a consequence, computing 
extensions, or 
determining wether a given formula belongs to one 
extension or all
 extensions can be answered without leaving the field of classical
 propositional logic. The 
transformation
 proposed is linear in the number of defaults.
\end{abstract}
\begin{flushleft}
\end{flushleft}
\section{INTRODUCTION}\label{sec01}
\subsection{MOTIVATION, 
CONTRIBUTION AND OUTLOOK}
Reiter's {\em default logic}~\cite{reiter80} is at once the most 
popular and the most controversial non-monotonic formalism dealing 
with uncertain information. Popular, because knowledge can be 
represented in a natural way in default logic. Controversial, because 
the concept of an extension as presented by Reiter is quite complex. 
Since default logic has been presented~\cite{reiter80} a lot of 
effort took place to establish relationships between default logic 
and other non-monotonic formalisms, such as {\em autoepistemic 
logic}~\cite{konolige88} and {\em 
circumscription}~\cite{lifschitz90}.

Poole~\cite{poole88} ({\em Theorist}) was --- to our knowledge --- 
the first one who 
tried
to embed Reiter's default logic in an {\em assumption-based 
framework}~\cite{dekleer86a,dekleer86b,FD93,keantsiknis92,keantsiknis93}.  
It
turned out that his proposition derives more 
extensions
than the original default theory allows.  The transformation proposed
by Ben-Eliyahu \& Dechter~\cite{dechter96} ({\em 
meta-interpretations}) results in a one-to-one correspondence of the 
original default theory and the their constructed propositional 
theory, but their transformation is in general is NP-complete.  
Another
quasi-transformation proposed by Bondarenko et al.\ 
\cite{bondarenko97} ({\em abstract argumentation-theoretic approach}) 
lacks a procedure to compute the extensions of the specified default
theory. In conclusion, the transformations mentioned so far have 
their weaknesses. That is the main reason, why we searched for a 
translation of propositional default theories into an 
assumption-based framework, called 
{\em argumentation 
systems}~\cite{km93,km95,kmah98,haenni98a,abhkl99}, to overcome  the 
weaknesses encountered so far.

In argumentation systems all 
inference is done without leaving the 
field
of classical propositional logic.  Informally speaking, argumentation
systems allow to judge open questions (hypotheses) about the unknown 
or
future world in the light of the given knowledge.  The problem is to 
determine a set of possible assumptions that allows to 
deduce
the hypothesis from the given knowledge. Note that the concept of 
argumentation systems has many
different meanings and we refer to \cite{cml98} for an overview.  For 
instance, the {\em argumentation framework} proposed by 
Prakken~\cite{prakken93}, which uses the same terminology as  in 
Bondarenko et al.\ 
\cite{bondarenko97}, is a derivation of Nute's~\cite{nute94} {\em 
defeasible logic}. So, Prakken's proposal has only the term {\em 
argumentation} with our approach in common and it is not an 
assumption-based framework. Therefore, we do not make a comparison 
between Prakken's~\cite{prakken93} and our approach. But it is worth 
noting that Prakken's proposal is close to 
Lukasiewicz's~\cite{lukasiewicz98} approach which extends still 
another derivation of Nute's defeasible logic with probabilities. 
Lukasiewicz proposal in turn has similar properties as the framework 
proposed by Benferhat et al.~\cite{benferhat2000}. However, in this
paper the notion of an argumentation system is defined in
Section~\ref{sec05}.

In this paper, we present a transformation of Reiter's default 
logic 
into
the framework of argumentation systems.  This 
transformation is
linear in the number of defaults and allows us to establish a 
bijection between extensions of the default theory and some subsets 
of assumptions of the argumentation system. Moreover, we provide 
procedures to answer any query on the specified default theory.
    
The rest of this paper is organized as follows: Section~\ref{sec012} 
repeats the necessary basic notions of propositional logic as a 
remainder. 
In Section~\ref{sec03} we briefly
present Reiter's default logic and introduce the transformation of 
propositional default theories into  associated propositional 
argumentation systems. 
Section~\ref{sec04} discusses the basic elements of propositional 
argumentation 
systems.  In
Section~\ref{sec05} we present the new results obtained using the 
transformation mentioned. We 
conclude with an outlook in Sections~\ref{sec06}. Proofs of the new 
theorems can be found in~\cite{dritan01b}.
    

\subsection{PRELIMINARIES}\label{sec012}
We assume that the reader is 
familiar with the basic concepts of propositional logic. We work with 
the usual propositional language $\ls_V$. That is, $\ls_V$ denotes 
set of all formulas which can be formed using the finite set of {\em 
propositions} $V = {v_1,...,v_n}$, where as usual, $\wedge$ denotes 
{\em conjunction}, $\vee$ {\em disjunction}, $\neg$ {\em negation} 
and $\rightarrow$ {\em implication}. A {\em literal} $\ell \in 
V^{\pm}$ is a proposition or the negation of a proposition. $\top$ 
denotes  {\em verum} and $\bot$ {\em falsum}. Lower-case Greek 
letters $\alpha, \beta, \ldots$ denote formulas and upper-case Greek 
letters $\Sigma, \Xi, \ldots$ denote sets of formulas called {\em 
theories}. For a set of formulas {\em satisfiability} and {\em 
consistency} is defined as usual. Propositional derivability is 
denoted by $\vdash$ and the corresponding consequence operator by 
$Th$. 

For a given a formula $\phi \in \ls_V$ and a subset of 
propositions $Q \subseteq V$, we are often  interested in computing a 
formula $\psi \in \ls_{Q}$ such that

\begin{quote}
\begin{itemize}
    \item[\rm (i)] $\phi \vdash \psi$ and
    \item[\rm (ii)] $\varphi \in \ls_{Q}$ and $\phi \vdash \varphi$ 
imply 
$\psi \vdash \varphi$.
\end{itemize}
\end{quote}
This is the {\em marginalization-problem}
\cite{khm99,hkl2000,hl2000,kmah98} which is a special case of
{\em literal forgetting}~\cite{reiter94} and equivalent to computing 
{\em prime implicates}~\cite{raimandekleer92}.  A 
formula~$\psi$
which satisfies the two conditions above is called a 
{\em marginal} of
$\phi$ with respect to $Q \subseteq V$, denoted by $\phi^{\downarrow 
Q}$. Note that $Th_{\ls_Q}(\phi^{\downarrow Q}) = Th_{\ls_V}(\phi) 
\cap \ls_{Q}$. We will not enter into the details how such a marginal 
can be computed. The interested reader is referred for example to 
Marquis~\cite{marquis2000}.

Finally, for any set $S$ we denote with 
$|S|$ the {\em cardinal number} of $S$, i.e. the number of elements 
contained in $S$, and with $2^S$ the {\em power set} of $S$, i.e. the 
set of all subsets of $S$.
\section{DEFAULT LOGIC AND TRANSLATION}\label{sec03}

\subsection{DEFAULT 
THEORIES AND EXTENSIONS}\label{sec031}
A Reiter's {\em default theory}~\cite{reiter80} is a pair $\dt$, 
where $\Sigma$ consists of
formulas over a propositional language $\ls$ and $\Delta$ is a set of
defaults. 
A {\em default} $\delta$ is a construct of the 
form\footnote{Note that different defaults have in general a 
different number of 
justifications; yet for the sake of simplicity, we do not index 
the~$k$'s 
in the sequel.}
    \begin{equation*}
	\frac{\alpha : \; \beta_{1} , \dots , \beta_{k}}{\gamma}, \; 
\text{where}
    \end{equation*}
    \begin{itemize}
	\item[\rm (i)] $\alpha, \beta_{1} , \dots , \beta_{k} , \gamma$ are 
formulas in 
	$\ls$,
	
	\item[\rm (ii)] $\pre(\delta) := \{\alpha\}$ is called the 
{\em prerequisite} of
	the default $\delta$,
	
	\item[\rm (iii)] $\jus(\delta) := \{\beta_{1} , \ldots , 
\beta_{k}\}$ is called 
the
	{\em justification} of the default $\delta$,
	
	\item[\rm (iv)] $\con(\delta) := \{\gamma\}$ is called the 
{\em consequence} of
	the default $\delta$.
    \end{itemize}
The selectors $\pre$, $\jus$ and $\con$ are naturally generalized to 
subsets $D \subseteq \Delta$ of defaults:
\begin{align*}
\pre(D) := 
\bigcup_{\delta \in D} \pre(\delta),\\
    \jus(D) := \bigcup_{\delta \in D} \jus(\delta),\\
    \con(D) := \bigcup_{\delta \in D} \con(\delta).
\end{align*}
Observe that the selectors $\pre$, $\jus$ and $\con$, respectively, 
return
sets of formulas of~$\ls$.  It is sometimes necessary to select the
formulas themselves.  Thus for any default $\delta$ with $k$
justifications, we define following selectors:
$$\p(\delta) := \alpha,\quad
    \j_{i}(\delta) := \beta_{i},\quad
    \c(\delta) := \gamma.$$
$\dt$ is called {\em finite} if both $\Sigma$ and $\Delta$ 
are finite.  In this paper we focus on finite default theories. 
Therefore let $P = \var(\dt)$ denote the set of all propositions 
occurring in the formulas of~$\dt$ and $\ls_P$ the propositional 
language of $\dt$.

The fundamental notion in default logic is the 
{\em extension}.  Intuitively, it 
denotes the deductive closure of a maximal set of formulas of $\dt$ 
containing $\Sigma$ and
``consistently'' adding the consequences of some defaults.
\begin{defn}\label{def33}
{\bf (Extension: \cite{reiter80}, Theorem 
2.1)}
    A set $E \subseteq \ls_{P}$ of formulas is called {\em extension} 
of a default 
theory $\dt$, where $P = \var(\dt)$,
    iff there exists a sequence $E_0, E_1, \ldots$ such that
    \begin{alignat*}{2}
	E_{0} &= \Sigma,&\; &{}\\
	E_{j+1} &= Th(E_{j}) \; \cup &\; \con(\{ &\delta \in \Delta 
:\p(\delta) \in E_{j} \text{ and }\\ 
&{} &\; &\neg\j_{i}(\delta) 
\notin E \text{ for
	every } i\}),\\
	E &= \bigcup_{j=0}^{\infty}E_{j}.&\; &{}
    \end{alignat*}
\end{defn}
When we write ``for every $i$'' ($\forall i$ for short), we mean for 
each 
justification of the
respective default.
\begin{exmp}\label{ex31}
    The default theory $\dt$ with $\Sigma = \{b \rightarrow \neg a 
\wedge \neg
    c\}$ and $\Delta = \{\frac{\top:a}{a} , \frac{\top:b}{b} ,
    \frac{\top:c}{c}\}$ has two extensions given by $E = Th(\Sigma 
\cup
    \{a,c\})$, and $E' = Th(\Sigma \cup \{b\})$.
\end{exmp}
Assume that $\Sigma$ of a given default theory $\dt$ is inconsistent. 
Then $Th(\Sigma)$ is the only extension of $\dt$ which is obviously 
inconsistent. Therefore, a 
default theory is called {\em consistent} iff it has at least one 
consistent extension. 
In this case $\Sigma$ must be 
consistent, since it is included in every 
extension. But consistency of the initial belief does not guarantee 
that the default theory has an extension.
\begin{exmp}
The default theory $\dt$ with $\Sigma = \emptyset$ and 
$\Delta =
    \{\frac{\top:p}{\neg p}\}$ has no extension!
\end{exmp}
Since we are not interested in trivial inconsistent extensions, we 
will say 
that a default theory has {\em no extension} iff the 
default 
theory $\dt$
lacks an extension or $\Sigma$ is inconsistent.

Default theories $\dt$ lacking an extension should not be confused 
with 
default
theories that have a unique consistent extension which is just the 
deductive 
closure of $\Sigma$.
\begin{exmp}\label{ex35}
    Let $\dt$ be a default theory with $\Sigma = \{p,q\}$ and $\Delta 
=
    \{\frac{p: \neg q}{\neg q}\}$.  The reader may verify that $E = 
    Th(\{p,q\})$ is the only extension.
\end{exmp}

Two concepts of consequence are of main interested when working with 
default theories:
\begin{defn}
    A formula $\phi$ is called {\em credulous consequence} of a 
default
    theory $\dt$ iff it belongs to at least one extension, and
    {\em skeptical consequence} iff it belongs to all extensions.
\end{defn}

\subsection{EMBEDDING GENERAL DEFAULT THEORIES}\label{sec022}
Let $\dt$ be a default theory with formulas in $\ls_P$, where $P =
\var(\dt)$. For any default $\delta \in \Delta$ consider the following
transformation $t$:
\begin{align}
    t(\delta) = &\{a_{\delta}^{\p} \rightarrow \neg \p(\delta)\} 
\;\cup\nonumber\\
    &\{a_{\delta,i}^{\j} \rightarrow \j_{i}(\delta): i 
=1,\ldots,|\jus(\delta)|\}\;
    \cup\nonumber\\ &\{a_{\delta}^{\c} \rightarrow \c(\delta)\}.
\end{align}
We call the triple $\langle a_{\delta}^{\p} , \{a_{\delta ,
i}^{\j}: i =1,\ldots,|\jus(\delta)|\} , a_{\delta}^{\c} \rangle$
{\em default assumption} and denote it by $a_{\delta}$. Moreover, 
let 
$A_\Delta = \{\langle a_{\delta}^{\p} , \{a_{\delta ,
i}^{\j}: i =1,\ldots,|\jus(\delta)|\} , a_{\delta}^{\c} \rangle: 
\delta \in \Delta\}$ denote the set of default assumptions. Often, we 
will
look at indexed defaults $\delta_{j}$ and abbreviate the default 
assumption
$a_{\delta_{j}}$ by $a_{j}$. For a default assumption 
$\langle 
a_{\delta}^{\p} , \{a_{\delta ,
i}^{\j}: i =1,\ldots,|\jus(\delta)|\} , a_{\delta}^{\c} \rangle \in 
A_\Delta$ we call $a_{\delta}^{\p}$ {\em prerequisitional}, 
$a_{\delta , 
i}^{\j}$
{\em justificational} ($\forall i$), and $a_{\delta}^{\c}$
{\em consequential assumption}. The sets 
$A_{\p} = \{a_{\delta}^{\p} 
: \delta\in\Delta\}$, 
$A_{\j}=\{a_{\delta,i}^{\j}: \delta\in\Delta, 
i=1,\ldots,|\jus(\delta)|\}$ and finally 
$A_{\c} = \{a_{\delta}^{\c} 
:\delta\in\Delta\}$ 
denote the corresponding sets of 
prerequisitional,
justificational, and consequential assumption, respectively. The set 

$A = A_{\p} \cup A_{\j} \cup A_{\c}$, which consists of pairwise 
distinct
propositions such that $A \cap P = \emptyset$, is called the set of 
{\em assumptions}. The pair
\begin{equation}
    \langle \Xi , A \rangle,\text{ were } \Xi =\Sigma \cup 
\bigcup_{\delta \in \Delta} t(\delta),
\end{equation}
is called the {\em (propositional) argumentation 
system} associated with the default theory $\dt$ (cf. 
Section~\ref{sec04}).
\begin{exmp}\label{ex51}
    Let $\dt = \langle \{e \vee o\} , \{\frac{e : r}{r}, \frac{o :
    r}{r}\}\rangle$ be a default theory, then the corresponding 
argumentation
    system is $\ag = \langle \{e \vee o , a_{1}^{\p} \rightarrow \neg 
e , 
a_{1}^{\j}
    \rightarrow r , a_{1}^{\c} \rightarrow r , a_{2}^{\p} \rightarrow 
\neg o ,
    a_{2}^{\j} \rightarrow r , a_{2}^{\c} \rightarrow r\}, 
\{a_1^{\p},a_2^{\p},a_1^{\j},a_2^{\j},a_1^{\c},a_2^{\c}\}\rangle$, 
where the 
set of
    default assumptions is $A_{\Delta} = \{\langle a_{1}^{\p}, 
\{a_{1}^{\j}\},
    a_{1}^{\c} \rangle , \langle a_{2}^{\p}, \{a_{2}^{\j}\}, 
a_{2}^{\c}
    \rangle\}$.
\end{exmp}

Any subset $\alpha \subseteq A_{\c}$ is called {\em consequential 
term}. Consequential terms   will allow us, using the framework of 
propositional argumentation systems, to decide wether a given default 
theory has extensions, characterize all extensions, and wether a 
given formula is a skeptical or credulous consequence of the default 
theory.

\section{PROPOSITIONAL 
ARGUMENTATION SYSTEMS}\label{sec04}
The major difference between 
argumentation systems~\cite{hkl2000,anrig00} and other 
non-mono\-to\-nic 
formalisms~\cite{reiter80,mccarthy80,moore85} is that in 
argumentation systems, all
the inference is done with classical propositional logic.
    
Consider a propositional theory $\Xi$ with formulas in $\ls_V$, where 
$V = \{v_{1}, v_{2}, \ldots, v_{m}\}$.  
From $V$ we choose a finite subset $A \subseteq V$ whose 
members are
called {\em assumptions}.  The other propositions in $P = V - A$ 
are
called {\em non-assumables}. For convenience, we enumerate the 
assumptions,
i.e.\ $A = \{a_{1}, \ldots , a_{n}\}$. The pair 
$\langle \Xi , A \rangle$ is
called a (propositional) {\em argumentation system} and $\Xi$ is 
said to
be an {\em argumentation theory}. So we 
consider a
fixed argumentation system $\langle \Xi , A \rangle$, with $\vdash$ 
as the derivability relation and $Th$ the corresponding closure 
operator. Note that the case where 
$A = \emptyset$, corresponds to the usual propositional
theory.

In an argumentation system $\langle \Xi , A \rangle$, the 
assumptions 
$A =
\{a_{1},\ldots,a_{n}\}$ of the argumentation theory $\Xi$ are 
essential for
expressing uncertain information.  They represent uncertain events, 
unknown
circumstances, or possible risks and outcomes. Formally, this 
is defined as follows:
    
\begin{defn}\label{def41}
    A {\em term} $\alpha$ is a subset of literals from $A^{\pm}$,  
so  that if $\ell \in \alpha$, then $\neg \ell \notin \alpha$. The 
set of all terms is denoted by $\CCA$.
\end{defn}
The non-assumables appearing in $\Xi$ are treated as 
classical
propositions for the given argumentation theory~$\Xi$.
\begin{defn}\label{def42}
     A term $\alpha \in \CCA$ is called
    {\em inconsistent} relative to $\Xi$ iff $\alpha \cup \Xi$ is 
unsatisfiable. $\ia(\Xi)$ denotes the set of inconsistent terms 
relative to $\Xi$. An inconsistent term $\alpha$ is called {\em 
minimal} iff no proper subset of $\alpha$ is inconsistent relative to 
$\Xi$. $\mu \ia(\Xi)$ denotes the set of all minimal inconsistent 
terms relative to $\Xi$, and is called the set of {\em minimal 
contradictions}. 
\end{defn}
In the sequel we will write $\alpha,\Xi \vdash \bot$ to indicate 
that~$\alpha$ 
is
inconsistent, and $\alpha,\Xi \nvdash \bot$ to indicate that~$\alpha$ 
is not 
inconsistent (relative to $\Xi$). 
$Th(\alpha,\Xi)$ denotes the deductive closure of $\alpha \cup \Xi$, 
and $\alpha,\Xi 
\vdash
\psi$ is an abbreviation for $\alpha \cup \Xi \vdash \psi$. With this 
notation the set of minimal contradictions can be represented by 
$\mu\ia(\Xi) = \mu\{\alpha \in \CCA : \alpha,\Xi \vdash \bot\}$.
\begin{exmp}\label{ex41}
    Let $\langle \Xi , A \rangle$ be an argumentation system, where 
$\Xi=\{a_1
    \rightarrow p , a_2 \rightarrow q , \neg p , \neg q\}$ is the
    argumentation theory and $A=\{a_1,a_2\}$ the set of assumptions.  
Then
    $\mu\ia(\Xi)=\{\{a_1\} , \{a_2\}\}$ is the set of minimal 
contradictions.
\end{exmp}
An argumentation system $\ag$ allows to reason 
about a
specified propositional formula $\phi \in \ls_V$, called {\em 
hypothesis}, with respect to $A$. The idea is to consider terms 
$\alpha$ which together with the argumentation theory $\Xi$ imply 
this hypothesis  $\phi$.
\begin{defn}\label{def43}{\bf 
(\cite{hkl2000}, Theorem 2.7)}
    Let $\langle \Xi , A \rangle$ be an argumentation system.  A 
term $\alpha$ is called a
    {\em supporting argument} for $\phi$ iff $\alpha , \Xi \vdash 
\phi$ and 
$\alpha' , \Xi \nvdash \bot$ for all $\alpha' \in \CCA$ such that 
$\alpha \subseteq \alpha'$.
\end{defn}
$\SP(\phi;\Xi)=\{\alpha\! \in\! \CCA: \alpha , \Xi \!\vdash\! \phi 
\text{ and } \alpha', \Xi\! \nvdash\! \bot  \text{ for all } \alpha' 
\in \CCA \text{ such that } \alpha \subseteq \alpha'\}$ denotes the 
corresponding set of supporting arguments.
A supporting argument $\alpha$ for a hypothesis $\phi$ 
describes an environment in which the hypothesis is a logical 
consequence of the given
argumentation theory, hence necessarily true. Moreover, extending 
this environment in such a way that $\alpha$ is still the case, does 
not lead to inconsistencies. For a general discussion of the 
requirement that $\alpha',\Xi \nvdash \bot$ for all $\alpha' \in 
\CCA$ such that $\alpha \subseteq \alpha'$ we refer to~\cite{KBH02}.

\begin{exmp}\label{ex42} 
    Let $\Xi=\{a_1 \rightarrow p , a_2 \rightarrow q , p 
\rightarrow
    \neg q\}$ and $A=\{a_1,a_2\}$.  The supporting arguments 
for~$p$ are
    $\SP(p;\Xi)=\{\{a_1 , \neg a_2\}\}$. Note that $\{a_1\},\Xi 
\vdash p$ and $\{a_1\},\Xi \nvdash \bot$, but $\{a_1,a_2\},\Xi \vdash 
\bot$, hence $\{a_1\}$ is not a supporting argument for $p$. 
Similarly, we get 
$\SP(q;\Xi)=\{\{\neg a_1 , 
a_2\}\}$.
\end{exmp}
\begin{defn}\label{def47}
    Let $\langle \Xi , A \rangle$ be an argumentation system, 
$\alpha$  a term, and $\ac \subseteq A^{\pm}$ a set of literals.  
The pair
    $\langle \alpha , \ac \rangle$ is called {\em structure} iff

\begin{quote}
    \begin{itemize}
	\item[\rm (i)] $\alpha , \Xi \nvdash \bot$, and
	
	\item[\rm (ii)] $\{\ell\} \cup \alpha , \Xi \nvdash \bot$, for every 
$\ell \in 
\ac$.
    \end{itemize}
\end{quote}
    $\alpha$ is called the {\em anchor} and $\ac$  the {\em set 
of
    irrelevant literals} of $\tstruct$ w.r.t.~$\Xi$.
\end{defn}
Note that $\{\ell\} \cup \alpha , \Xi \nvdash \bot \text{ for every } 
\ell
\in \ac$ does not guarantee that $\{\ell_{i} , \ell_{j}\} \cup \alpha 
, \Xi
\nvdash \bot$ for $\ell_{i},\ell_{j} \in \ac$ and $i \neq j$. The 
utility of structures will become clear in Section~\ref{sec05}.
    
We close our discussion of terms and scenarios and refer
to~\cite{hkl2000,km95,kmah98,km93,abhkl99,haenni98a,hl2000} for 
readers 
wishing
more details concerning this issue.  For computational purposes, 
especially
computing minimal contradictions and approximation techniques see
\cite{hkl2000,haenni01b}.
\section{EXTENSIONS AND 
TERMS}\label{sec05}
Here, we characterize extensions in the 
corresponding propositional argumentation system using 
structures~$\tstruct$
(cf.\ Definition~\ref{def47}), and provide, based on the anchor of 
some structures, an inductive procedure to compute extensions, if the 
original default theory has any. Here, only consequential terms (cf. 
Section~\ref{sec022}) are 
allowed
as anchor $\alpha$, i.e.\ $\alpha \subseteq A_{\c}$ and the set of
irrelevant literals $\ac$ consists only of justificational
assumptions, i.e.\ $\ac \subseteq A_{\j}$.  What follows is close 
in spirit to Lukaszewicz's~\cite{lukaszewicz91} semantics for 
Reiter's
default logic.
    
A pair $\langle \alpha , \ac \rangle$, where $\alpha \subseteq 
A_{\c}$
is a consequential term and $\ac \subseteq A_{\j}$ is a set of
justificational assumptions is called 
{\em consequence-justification
pair}, {\em CJ-pair} for short.
\begin{defn}\label{def51}
    Let $\dt$ be default theory, $\langle \Xi , A \rangle$ its 
corresponding
    argumentation system, $\alpha \subseteq A_{\c}$ a 
consequential term
    and $a = \das$ a default assumption.  We say that $a$ is
    {\em applicable} with respect to $\alpha$ iff
\begin{quote}
    \begin{itemize}
	\item[\rm (i)] $\{a^{\p}\} \cup \alpha \in \ia(\Xi)$ and
	
	\item[\rm (ii)] $\{a_{i}^{\j}\} \cup \alpha \notin \ia(\Xi)$ for $i =
	1,\ldots,k$.
    \end{itemize}
\end{quote}
\end{defn}
Note that we do not check if $\{a^{\c}\} \cup \alpha , \Xi \vdash
\bot$.  The idea is to obtain a sequence of consequences to add 
successively to the consequential term according to 
Definition~\ref{def33} in 
an argumentation system.    
\begin{defn}\label{def52}
    To each default assumption $a = \das$ we assign a mapping, 
denoted by $a^{\prec}$, from CJ-pairs into CJ-pairs specified by

\begin{quote}
\begin{itemize}
\item[\rm (i)] $a^{\prec}(\langle 
\alpha , \ac \rangle) =
\langle \{a^{\c}\} \cup \alpha , 
\{a_{1}^{\j},\ldots,
	    a_{k}^{\j}\} \cup \ac \rangle$, if $\langle \alpha , 
	    \ac \rangle$ is a structure and $a$ is applicable w.r.t. 
$\alpha$;
\item[\rm (ii)] $a^{\prec}(\langle \alpha , \ac \rangle) = 
\tstruct$,
if $\langle \alpha , 
	    \ac \rangle$ is a structure and
	    $a$ is not applicable w.r.t. $\alpha$;
\item[\rm (iiii)] 
$a^{\prec}(\langle \alpha , \ac \rangle) = \langle \{\bot\} , 
\emptyset \rangle$, otherwise. 
\end{itemize}
\end{quote}
\end{defn}
If a default assumption is applicable with respect to some 
consequential
term, then we have to include its consequential assumption to the 
already
given consequential term.  But this inclusion may lead to
inconsistencies. Consequently, we have to check if the obtained 
CJ--pair 
is still a structure.
\begin{exmp}\label{ex52}
    Consider $\langle \{p\},\{\frac{p : q}{ \neg q}\} \rangle$ with 
$\Xi = \{p
    , a^{\p} \rightarrow \neg p , a^{\j} \rightarrow q , a^{\c} 
\rightarrow
    \neg q\}$.  If we take the structure $\langle \{\top\} , \emptyset
    \rangle$, then $a = \langle a^{\p} , \{a^{\j}\} , a^{\c} \rangle$ 
is 
applicable
    w.r.t.\ $\{\top\}$.  Thus $a^{\prec}(\langle \{\top\} , \emptyset 
\rangle)
    = \langle \{a^{\c}\}, \{a^{\j}\} \rangle$, but $\{a^{\j}\} \cup 
\{a^{\c}\} ,
    \Xi \vdash \bot$, hence $\langle \{a^{\c}\} , \{a^{\j}\} \rangle$ 
is not a
    structure.
\end{exmp}
\begin{defn}\label{def54}
    A structure $\tstruct$ is called {\em accessible} (with 
respect to 
    $A_{\Delta}$) iff there is a sequence $\tseq = \langle a_{1} , 
    a_{2} , \ldots , a_{q}\rangle$ of non--repeating default 
assumptions 
    such that 
\begin{quote}
    \begin{itemize}
	\item[\rm (i)] $\langle \alpha_{0} , \ac_{0} \rangle = \langle 
\{\top\} , 
	\emptyset\rangle$;
	
	\item[\rm (ii)] $a_{j}$ is applicable w.r.t.\ $\alpha_{j-1}$ and 
	$\langle \alpha_{j} , \ac_{j} \rangle = a_{j}^{\prec}(\langle 
	\alpha_{j-1},\ac_{j-1}\rangle)$ for $j = 1, 2, \ldots,q$;
	
    \item[\rm (iii)] $\langle \alpha_q,\ac_q\rangle = 
\langle\alpha,\ac\rangle$

	\item[\rm (iv)] there is no default assumption $a_{i} \in 
A_{\Delta}$ which is 
	applicable w.r.t.\ $\alpha$ but not included in $\tseq$.
    \end{itemize}
\end{quote}
    Such a $\tseq$ is called a {\em generating sequence} of 
$\tstruct$.
\end{defn}
If $\tseq$ is a generating sequence of a structure 
$\tstruct$, then $\alpha$ contains all consequential assumptions of 
the default assumptions appearing in $\tseq$, $\ac$ contains the 
union of all sets of justificational assumptions of the default 
assumptions appearing in $\tseq$, and $\tstruct$ is still a 
structure. In particular $\alpha \subseteq A_{\c}$.
\begin{exmp}
    Consider the default theory 
    $\langle \emptyset,\{\frac{\top : p}{p} , 
    \frac{\top : q}{q}\} \rangle$ with corresponding 
    $\Xi = \{\neg a_{1}^{\p} , a_{1}^{\j} \rightarrow p , 
    a_{1}^{\c} \rightarrow p , \neg a_{2}^{\p} , a_{2}^{\j} 
    \rightarrow q , 
    a_{2}^{\c} \rightarrow q\}$. 
    If we take the structure $\langle \{\top\} , \emptyset
    \rangle$, then $\langle a_{1} , a_{2} \rangle$ and $\langle 
    a_{2} , a_{1} \rangle$ are generating sequences of the same 
    structure $\langle 
    \{a_{1}^{\c},a_{2}^{\c}\} , \{a_{1}^{\j},a_{2}^{\j}\}\rangle$.
\end{exmp}
Obviously, there may be different generating sequences for the same
structure. But two generating sequences of the same structure contain 
the same default assumptions, i.e.\ one is a permutation of the 
other~\cite{dritan01b}.

The definition of accessible structures is justified by the next 
theorem, which can be regarded as the dual result of Theorem 5.63 
in~\cite{lukaszewicz91}.  
Duality in the sense that Lukaszewicz'~\cite{lukaszewicz91} theorem 
is 
a semantic characterization of extensions and our result is a 
syntactic characterization of extensions. In
fact the proof follows Lukaszewicz~\cite{lukaszewicz91}.
\begin{thm}\label{th52}{\bf \cite{dritan01b}}
    Let $\dt$ be a default theory and $\langle \Xi , A \rangle$ the
    corresponding argumentation system.  There is a bijection between
    accessible structures of $\langle \Xi , A \rangle$ and 
    consistent extensions of $\dt$.
\end{thm}
The theorem above states only that an accessible structure 
contains the information
concerning some corresponding extension of the given default theory 
$\dt$.  It says
nothing about how the extension can be 
characterized.  This observation motivates our next definition to 
simplify
terminology.
\begin{defn}\label{def56}
    The anchor $\alpha$ of an accessible structure 
    $\tstruct$ is called {\em default
    term}.  The set of all default terms is denoted by 
$\mathrm{dt}(\Xi)$.
\end{defn}
The notion of a default term permits to derive some corollaries of 
Theorem~\ref{th52}.
\begin{cor}\label{cor51}{\bf \cite{dritan01b}}
    Let $\dt$ be a default theory with $P = \var(\dt)$ and 
    $\langle \Xi , A \rangle$ the
    corresponding argumentation system.  
\begin{quote}

\begin{itemize}
\item[\rm (i)] $\Sigma$ is inconsistent iff 
$\mu\ia(\Xi) = \{\{\top\}\}$.
\item[\rm (ii)] $\dt$ has no extension 
iff $\mathrm{dt}(\Xi)= \emptyset$ and $\mu\ia(\Xi) \neq 
\{\{\top\}\}$.
\item[\rm (iii)] $E = Th(\Sigma)$ is the only 
extension iff $\mathrm{dt}(\Xi)=\{\top\}$.
\item[\rm (iv)] $E = 
Th((\alpha,\Xi)^{\downarrow P})$ is an extension of $\dt$ iff 
$\alpha\in\mathrm{dt}(\Xi)$.
\end{itemize}
\end{quote}
\end{cor}
Now, that we have a complete characterization of extensions for a 
given default theory $\dt$ in terms of the default terms 
$\mathrm{dt}(\Xi) \subseteq 2^{A_{\c}}$ of the corresponding 
argumentation system $\langle \Xi , A \rangle$, it remains to 
determine how those default terms can be computed. Interestingly, all 
we need to do that are the minimal contradictions $\mu\ia(\Xi)$.  The 
minimal contradictions $\mu\ia(\Xi)$ are known to play a central role 
in argumentation systems, for example for computing supporting 
scenarios~\cite{hkl2000} and in model-based diagnostics and 
reliability~\cite{AK02}. Therefore a lot of effort has been made to 
develop efficient algorithms~\cite{hkl2000,haenni2001b} to compute 
$\mu\ia(\Xi)$. So default logic can profit from those efforts.
\begin{thm}\label{th53}{\bf \cite{dritan01b}}
    Let $\dt$ be a default theory and
    $\langle \Xi , A \rangle$ the corresponding argumentation 
system.  A term $\alpha \subseteq A_{\c}$ different from $\{\top\}$ 
is a default 
term iff
    there is an enumeration $a_{1}^{\c} , \ldots , a_{q}^{\c}$ of
    the consequential assumptions $a_{j}^{\c} \in \alpha$ such that

\begin{quote}
    \begin{itemize}
	\item[\rm (i)] $\{a_{1}^{\p}\} \in \ia(\Xi)$,
	
	\item[\rm (ii)] For every $j = 1,\ldots,q-1$ we have 
$\{a_{j+1}^{\p}\} \cup \{a_{1}^{\c} , \ldots , a_{j}^{\c}\} \in 
\ia(\Xi)$,
	
	\item[\rm (iii)] For every $j = 1,\ldots,q$ and every~$i$ we have 
$\{a_{ji}^{\j}\} \cup \alpha \notin \ia(\Xi)$,
	
	\item[\rm (iv)] For every $a_{l}^{\p} \in (A_{\p} -
	\{a_{1}^{\p},\ldots,a_{q}^{\p}\})$ we have
	\begin{itemize}
	    \item[\rm (a)] $\{a_{l}^{\p}\}
	    \cup \alpha \notin \ia(\Xi)$, \underline{{\em or}}
	    
	    \item[\rm (b)] $\{a_{li}^{\j}\}
	    \cup \alpha \in \ia(\Xi)$ for some~$i$.
	\end{itemize}
    \end{itemize}
\end{quote}
\end{thm}
Note that if $\alpha \in \mu\ia(\Xi)$, then any term 
$\alpha'$, such that $\alpha \subseteq \alpha'$, is inconsistent, 
i.e. $\alpha' \in \ia(\Xi)$. Clearly, only the minimal contradictions 
$\mu\ia(\Xi)$ will be computed. 

Theorem~\ref{th53} indicates an iterative 
procedure for computing default terms given the translation of the 
default theory and the respective minimal contradictions. The 
following examples shows how. 
\begin{exmp}
We consider for each case 
of Corollary~\ref{cor51} a small example.
\begin{itemize}
\item For 
$\dt = \langle \{p, \neg p\}, \{\frac{p: q}{q}\}\rangle$, $t(\dt) = 
\langle \{p,\neg p,a^{\p} \rightarrow \neg p, a^{\j} \rightarrow q, 
a^{\c} \rightarrow q\}, \{a^{\p},a^{\j},a^{\c}\}\rangle$ and 
$\mu\ia(\Xi) = \{\{\top\}\}$. Clearly, $\Sigma \vdash \bot$.
\item 
For $\dt = \langle \emptyset, \{\frac{\top: \neg p}{p}\}\rangle$ we 
have $t(\dt) = \langle \{\neg a^{\p}, a^{\j} \rightarrow \neg p, 
a^{\c} \rightarrow p\}, \{a^{\p},a^{\j},a^{\c}\}\rangle$ and 
$\mu\ia(\Xi) = \{\{a^{\p}\},\{a^{\c},a^{\j}\}\}$. The only 
potentially possible default term different from $\{\top\}$ is 
$\{a^{\c}\}$. $\{a^{\p}\} \in \mu\ia(\Xi)$, so point (i) of 
Theorem~\ref{th53} is fulfilled. But $\{a^{\c},a^{\j}\} \in 
\mu\ia(\Xi)$, thus (iii) of Theorem~\ref{th53} is violated, thus 
$\{a^{\c}\}$ is no default term. Hence we conclude that there is no 
default term and $\dt$ has no extension.
\item For $\dt = \langle \{e 
\vee o\} , \{\frac{e : r}{r}\}\rangle$ we have $t(\dt) = \langle \{e 
\vee o , 
a^{\p} \rightarrow \neg e , a^{\j} \rightarrow r , a^{\c} 
\rightarrow r\} ,\{a^{\p}, a^{\j}, a^{\c}\} \rangle$ and $\mu\ia(\Xi) 
= \{\{\bot\}\}$. As $\{a^{\p}\} \notin \mu\ia(\Xi)$ (point (i)), 
there is no default term different from $\{\top\}$ Indeed, this 
default theory has a single extension given by $E = Th(\{e \vee 
o\})$.
\item Finally, for $\dt = \langle \emptyset , \{\frac{\top : 
c}{\neg d} , 
\frac{\top
    : d}{\neg e} , \frac{\top : e}{\neg f}\} \rangle$ we have  $\Xi = 
\{\neg
    a_{1}^{\p} , a_{1}^{\j} \rightarrow c , a_{1}^{\c} \rightarrow 
\neg d ,
    \neg a_{2}^{\p} , a_{2}^{\j} \rightarrow d , a_{2}^{\c} 
\rightarrow \neg e
    , \neg a_{3}^{\p} , a_{3}^{\j} \rightarrow e , a_{3}^{\c} 
\rightarrow \neg
    f\}$ and $\mu \ia(\Xi) 
=
    \{\{a_{1}^{\p}\}, \{a_{2}^{\p}\}, \{a_{3}^{\p}\}, \{a_{1}^{\c} ,
    a_{2}^{\j}\}, \{a_{2}^{\c} , a_{3}^{\j}\}\}$. There are three 
    prerequisitional assumptions to start with:  
    \begin{itemize}
	\item First we start with $\{a_{1}^{\p}\}$. The hypothesis that 
$\{a_{1}^{\c}\}$ is a default term is rejected, since $\{a_{3}^{\p}\} 
\in \mu\ia(\Xi)$(point (a) of (iv) is violated). Thus 
$\{a_{1}^{\c},a_{3}^{\c}\}$ could be a default term. And indeed 
$\{a_{1}^{\c},a_{3}^{\c}\}$ is a default term, as (i) to (iv) is 
satisfied.
	\item Second we start with $\{a_{2}^{\p}\}$. The hypothesis that 
$\{a_{2}^{\c}\}$ is a default term is rejected, since $\{a_{1}^{\p}\} 
\in \mu\ia(\Xi)$(point (a) of (iv) is violated). Thus 
$\{a_{2}^{\c},a_{1}^{\c}\}$ could be a default term. But 
$\{a_{2}^{\c},a_{1}^{\c}\}$ being a default term must be rejected, 
since $\{a_{1}^{\c},a_{2}^{\j}\} \in \mu\ia(\Xi)$ (point (iii) is 
violated).
	\item Finally we start with $\{a_{3}^{\p}\}$. We already know that 
$\{a_{1}^{\c},a_{3}^{\c}\}$ is a default term. Thus we have to test 
if $\{a_{3}^{\c},a_{2}^{\c}\}$ is a default term, as $\{a_{2}^{\p}\} 
\in \mu\ia(\Xi)$ ($\{a_{3}^{\c}\}$ being a default term violates 
point (a) of (iv)). Since $\{a_{2}^{\c} ,
    a_{3}^{\j}\} \in \mu\ia(\Xi)$, the hypothesis 
$\{a_{3}^{\c},a_{2}^{\c}\}$ being a default term is rejected (point 
(iii) is violated).
    \end{itemize}
    So, $Th(\{\neg d, \neg f\})$ is the only 
extension of
    this default theory.
\end{itemize}
\end{exmp}

We need some notation to answer the question which formulas $\phi 
\in \ls_{P}$ are provable in a
default theory $\dt$, where $P = \var(\dt)$.  For a supporting 
argument $\alpha \in \SP(\phi;\Xi)$ 
we define $\alpha^+ = \alpha \cap A$ (i.e. we forget all literals 
$\ell \in \alpha$ such that $\ell = \neg a_i$ and $a_i \in A$).
\begin{thm}\label{th54}{\bf \cite{dritan01b}}
    Let $\dt$ be a default theory, $\langle \Xi , A \rangle$ the 
corresponding
    argumentation system, and $\phi \in \ls_{P}$ a formula, where $P =
    \var(\dt)$.  
\begin{quote}
    \begin{itemize}
	\item[\rm (i)] $\phi$ is a credulous consequence of $\dt$ iff 
$\{\alpha^+: \alpha \in \SP(\phi;\Xi)\} \cap
	\mathrm{dt}(\Xi) \neq \emptyset$;
	
	\item[\rm (ii)] $\phi$ is a skeptical consequence of $\dt$ iff 
$\mathrm{dt}(\Xi) 
\subseteq
	\{\alpha^+: \alpha \in \SP(\phi;\Xi)\}$.
    \end{itemize}
\end{quote}
\end{thm}
\begin{exmp}
    The default theory $\dt$, where $\Sigma = \{b \rightarrow d, c 
\rightarrow
    d\}$ and $\Delta = \{\frac{\top : \neg c}{b} , \frac{\top : \neg 
b}{c}\}$
    has two extensions $E_{1} = Th(\{b,d\})$ and $E_{2} = 
Th(\{c,d\})$. 
    Clearly, $b$ and $c$ are credulous consequences of $\dt$, whereas 
$b \vee
    c$ and $d$ are skeptical consequences.  The translation returns 
$\langle
    \Xi , A \rangle$ with $\Xi = \{b \rightarrow d, c \rightarrow d, 
\neg
    a_{1}^{\p}, \neg a_{2}^{\p}, a_{1}^{\j} \rightarrow \neg c, 
a_{2}^{\j}
    \rightarrow \neg b , a_{1}^{\c} \rightarrow b, a_{2}^{\c} 
\rightarrow c\}$. 
    One can verify that $\mathrm{dt}(\Xi) = 
\{\{a_{1}^{\c}\},\{a_{2}^{\c}\}\}$, $\{\alpha^+: \alpha \in 
\SP(b;\Xi)\} = \{\{a_{1}^{\c}\}\}$, 
$\{\alpha^+: \alpha \in 
\SP(c;\Xi)\} = \{\{a_{2}^{\c}\}\}$  and 
$\{\alpha^+: \alpha \in 
\SP(b \vee c;\Xi)\} = \{\alpha^+: \alpha \in \SP(d;\Xi)\} = 
\{\{a_{1}^{\c}\},\{a_{2}^{\c}\}\}$.
\end{exmp}
\section{CONCLUSION}\label{sec06}
We have provided a translation from default theories into 
argumentation 
systems which is linear in the number of defaults, from
which we can decide if the given default theory has no extensions 
and  if it has consistent extensions. Moreover, we are able to 
characterize all extensions, without leaving the monotone framework 
of propositional logic. All we need are
assumptions. Finally, for any given formula we can decide if it is a 
skeptical or credulous consequence of the original default 
theory.
    
\section*{Acknowledgments}

This work has been supported by grant No.\ 2000-061454.00 of the Swiss
National Foundation for Research. 

We would like to thank Philippe Besnard for his useful comments on 
earlier drafts of this article. We owe special thanks to the referees 
for their constructive and valuable suggestions which contributed to 
the final form of this article.


\end{document}